\documentclass[letterpaper]{article}
\usepackage{natbib,alifeconf,algorithm,algorithmic,mathtools,amsmath}

\title{Coevolutionary Neural Population Models}
\author{Nick Moran$^{1}$ \and Jordan Pollack$^{1}$ \\
\mbox{}\\
$^1$Brandeis University, 415 South St, Waltham, MA 02453 \\
nemtiax@brandeis.edu \\
pollack@brandeis.edu}

\begin{document}
\maketitle

\begin{abstract}
We present a method for using neural networks to model evolutionary population dynamics, and draw parallels to recent deep learning advancements in which adversarially-trained neural networks engage in coevolutionary interactions.  We conduct experiments which demonstrate that models from evolutionary game theory are capable of describing the behavior of these neural population systems.
\end{abstract}

\section{Introduction}

Recent works in the field of deep learning have developed algorithms which train neural networks through interaction, rather than through optimization of a static loss function.  These algorithms display dynamics which are similar to those observed in evolutionary simulations and those described by evolutionary game theory. One such algorithm, Generative Adversarial Networks (GANs) by \citep{goodfellow2014generative}, seeks to train a network to generate realistic images based on a training set.  It uses a pair of networks trained against one another to simultaneously develop a network capable of generating images and a discriminator network capable of evaluating images for realism.  Of particular interest to this work is the design of the generator network, which takes as input vectors sampled from a random distribution, and transforms them into output images.  Through the lens of evolutionary simulation, the output distribution induced by the generator can be viewed as a population of candidate images which are evaluated for fitness by the discriminator.  The members of this population compete with each other to occupy niches of realistic image types, and cooperate to collectively match the distribution in the training set.

\begin{figure}
\centering

\includegraphics[scale=0.33]{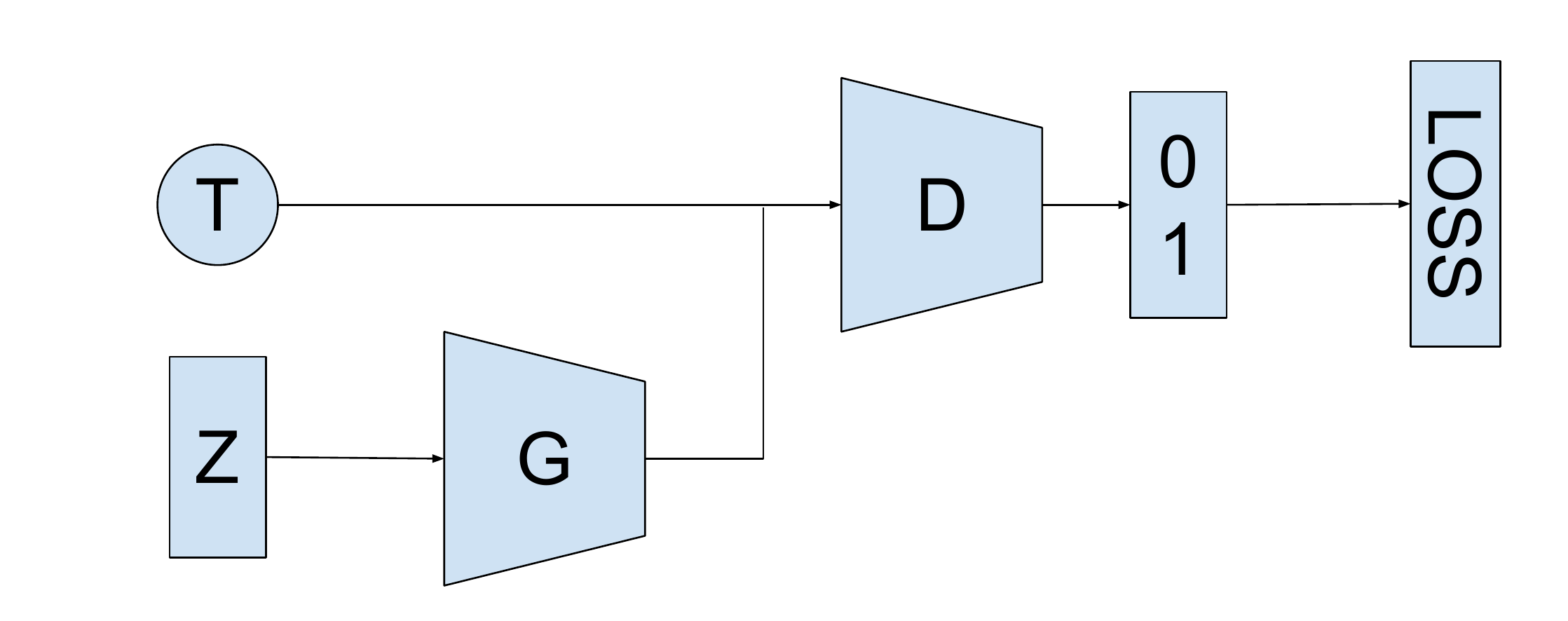}

\caption{The layout of a standard GAN.  $G$ and $D$ are the generator and discriminator networks, $T$ is the set of training examples, $Z$ is the latent input of $G$.}
\label{fig:gan_diagram}
\end{figure}

We will investigate the following questions: Does embedding evolutionary population dynamics in the substrate of a neural network trained by stochastic gradient descent fundamentally alter the behavior of the system? Can existing analysis tools from the fields of evolutionary computation and artificial life predict the behavior of such systems, or explain divergence from the standard predictions? To this end, we first present a neural model to simulate a coevolutionary population over time. We then conduct experiments using this model in which we simulate well-understood matrix games, and compare the resulting behavior to that obtained from traditional simulation techniques. We thereby lay the groundwork for future investigations into the application of evolutionary game theory to more complex adversarially-trained neural networks (such as GAN models), as well as the use of such models as a testbed for artificial life simulations.


\section{Background}

There have been many intersections between the fields of evolutionary computation and neural networks.  Neuro-evolution algorithms such as GNARL \citep{angeline1994evolutionary}, NEAT \citep{stanley2002evolving} and HyperNEAT \citep{stanley2009hypercube} employ evolutionary algorithms to evolve the structure and weights of neural networks.  \citep{miikkulainen2017evolving} proposes a hybrid approach in which network structure is evolved while weights are trained via back-propagation.  

Neural networks have also been used as controllers for artificial agents in evolutionary simulations, as proposed in \citep{harvey1992issues} and \citep{sims1994evolving}.  Recurrent Neural Networks are evolved to control physically embodied robots in \citep{lipson2000automatic}.  

Deep reinforcement learning has applied neural models to multi-agent systems.  \citep{bloembergen2015evolutionary} gives an overview of the applicability of evolutionary models to reinforcement learning in general.  In \citep{such2017deep}, genetic algorithms are proposed as method to evolve deep neural networks in reinforcement learning settings.

\subsection{Generative Adversarial Networks}

We present here a brief overview of the design of GAN systems as described in \citep{goodfellow2014generative}.  Figure~\ref{fig:gan_diagram} shows the basic layout of a GAN model.  Two neural networks, $G$ (the generator) and $D$ (the discriminator) are trained adversarially, with the goal of training $G$ to produce novel realistic images.  $G$ takes as input a random vector drawn from a random distribution, $Z$.  It transforms that vector into an output image, which is then passed to $D$. $D$ takes as input images either drawn from a training set of real images, $T$, or the output of $G$, and attempts to classify them as either real ($0$) or fake ($1$).  $D$ is trained by minimizing the error of its classifications through gradient descent.  $G$, on the other hand, is trained by maximizing the error of $D$ on $G$'s outputs.  

\begin{figure}
\centering

\includegraphics[scale=0.33]{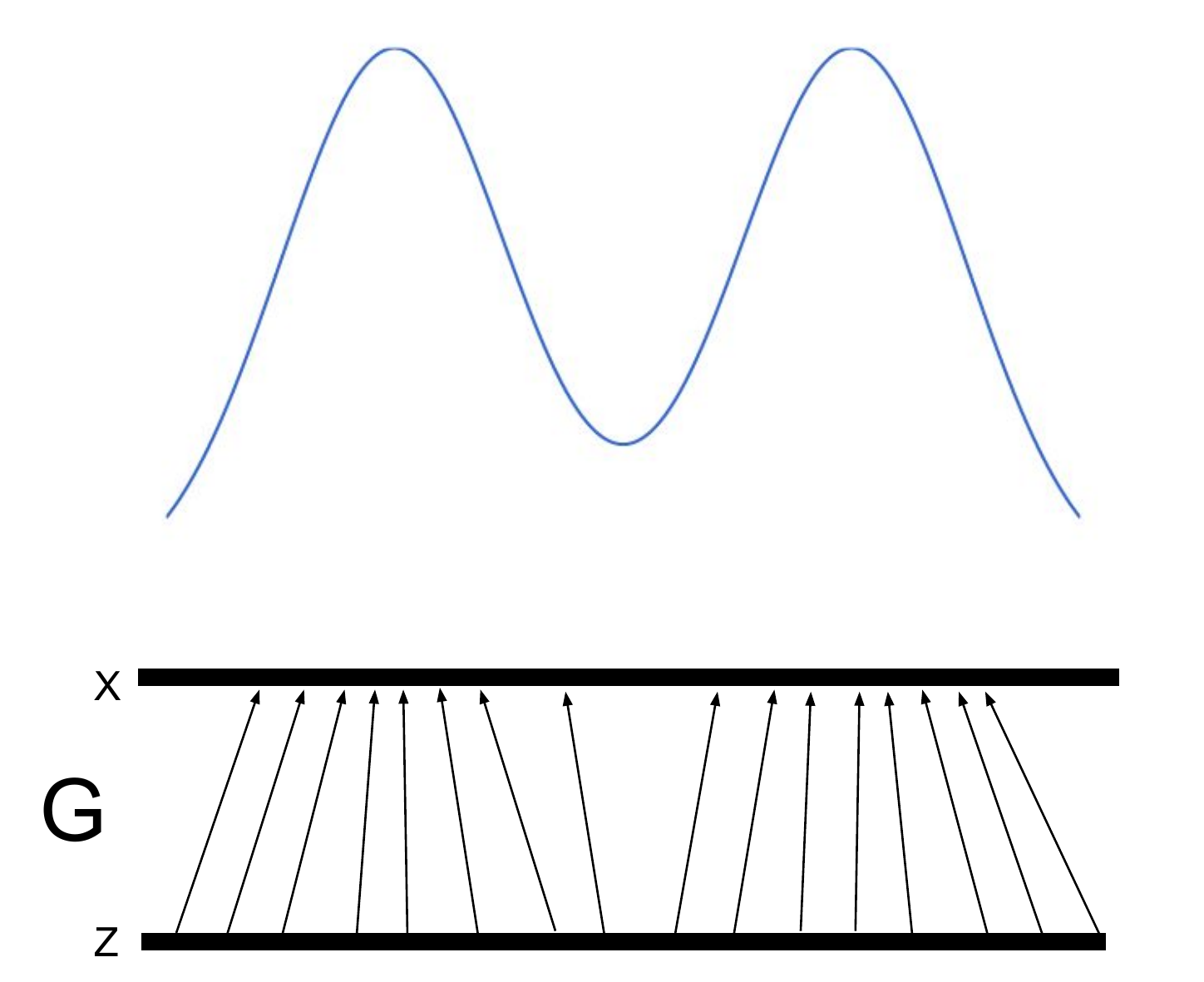}

\caption{Transformation of a simple uniform distribution, $Z$, to a more complex bimodal distribution, $X$, by an example generator, $G$.}
\label{fig:GeneratorBimodal}
\end{figure}

We are particularly concerned with the architecture of $G$, as it will form the basis for our neural population model.  $G$ transforms a simple distribution, $Z$ into a complex distribution (an approximation of the manifold of realistic images from which the samples in $T$ are drawn).  In order for $G$ to successfully fool $D$, it must not only produce realistic samples, it must produce them with the proper frequencies.  Figure~\ref{fig:GeneratorBimodal} shows a simplified example in which a generator transforms samples from a uniform distribution to samples from a bimodal distribution.  In an actual GAN, the output distribution is a high-dimensional distribution over the space of possible images of some size.

\subsection{Matrix Games and Evolutionary Game Theory}
Matrix games have been used as a testbed for analyzing evolutionary dynamics in computer models as presented in \citep{smith1973logic}.  This analysis is often focused on the discovery of evolutionarily stable strategies (ESS) \citep{smith1972game}, and the dynamics leading towards them. \citep{fogel1998instability} and \citep{ficici2000effects} examine the ability of evolutionary models to discover and maintain an ESS using the Hawk-Dove game.

\section{Neural Population Models}

Our model is concerned with the class of matrix games defined by a fixed set of strategies $S$ and a payoff matrix $M$ of dimension $|S| \times |S|$ which defines the payoff received by each strategy when played against each other strategy.  We will use the notation $M(s_1,s_2)$ to indicate the payoff received by strategy $s_1$ against $s_2$, where $s_1$ and $s_2$ may be pure or mixed strategies.  If they are mixed strategies, $M(s_1,s_2)$ is the weighted average of payoffs received.

The goal of our model is to represent the distribution of the set of strategies in a population, which we represent as $P$.  Instead of explicitly representing that distribution, we will construct a function $F_\theta$ (in the form of a neural network with weights $\theta$) which transforms a simple input distribution $Z$ to $P$.  In this work, we let $Z$ be a uniform distribution over $[0,1]^n$ for small $n$.

Given a sample $z$ drawn from $Z$, we can compute $F_\theta(z)$ to determine the strategy associated with $z$.  For example, in a game with three possible strategies, $A$, $B$, and $C$, we might have $F_\theta(z)=[0.3,0.5,0.2]$, indicating a mixed strategy composed of $30\% A$, $50\% B$ and $20\% C$.  To determine the overall composition of the population, we repeatedly sample from $Z$ and apply $F_\theta$ to retrieve a representative sample of $P$.

\subsection{Model Architecture}

We represent $F_\theta(z)$ as a feed-forward neural network in which the input is passed through a fully-connected layer containing ten hidden units with sigmoid activations, followed by a second fully-connected layer with $|S|$ output units and a softmax activation.  The softmax operation in the output layer has the effect of normalizing the output values to sum to one, so as to represent a valid mixed strategy in the game.  The number of hidden units as well as the depth of the network are effectively hyperparameters of the model.  More complex games with many strategies may require more complex networks to successfully model.

\subsection{Training}

To model the evolutionary trajectory of the population, we use stochastic gradient descent (SGD) to optimize the weights $\theta$ to maximize the payoffs individual samples from the population receive when matched against opponents also drawn from the population.  At each step of training, we draw two sets of random samples from $Z$, called $z_1$ and $z_2$, which will serve as mini-batches for the SGD algorithm.  $z_1$ and $z_2$ will have dimension $b \times n$, where $b$ is the mini-batch size and $n$ is the dimension of $Z$.  We then compute $p_1 = F_\theta(z_1)$ and $p_2 = F_\theta(z_2)$, which have dimension $b \times s$ where $s$ is the number of strategies in the game.  We then compute fitness values $M(p_1,p_2)$ for each row of $p_1$ by competing it against the corresponding entry of $p_2$ using the game matrix.

We now have $M(F_\theta(z_1),F_\theta(z_2))$, which is a differentiable function parameterized by $\theta$.  We can therefore compute gradients for $\theta$, and apply gradient ascent to maximize the payoff received.  Critically, we do not treat the computation of $F_\theta(z_2)$ as a differentiable component of the system.  Instead, the values of $F_\theta(z_2)$ are treated as constants.  This is done to prevent the system from improving the payoff of $F_\theta(z_1)$ by moving $F_\theta(z_2)$ towards a worse strategy.

\begin{algorithm}                     
\caption{Training Procedure}          
\label{alg1}                           
\begin{algorithmic}
    \REQUIRE $b$, the batch size, $n$, the dimension of $Z$
    \FOR{number of training iterations}
        \STATE Draw samples $z_1$ and $z_2 \in [0,1]^{b \times n}$ from $Z$
        \STATE Compute $F_\theta(z_1)$ and $F_\theta(z_2)$
        \STATE Compute $M(F_\theta(z_1),F_\theta(z_2))$
        \STATE Compute $\nabla_\theta \frac{1}{b}\sum{M(F_\theta(z_1),F_\theta(z_2))}$, treating $F_\theta(z_2)$ as a constant
        \STATE Update $\theta$ by ascending $\nabla_\theta$
    \ENDFOR
\end{algorithmic}
\end{algorithm}

\subsection{Initialization}

In most neural network applications, the parameters of the network are initialized to small random values.  This means that the initial outputs of the network are unpredictable.  However, in many evolutionary applications it is desirable to be able to initialize the model at a variety of starting configurations to observe the different trajectories that result.  For example, in the Hawk-Dove-Retaliator game, Maynard Smith found two attractors, such that the fate of the population depends on the initial proportions of the three strategies.

We propose an optional period of non-adversarial training to initialize the network to output a desired starting configuration.  Given a target distribution $D$, which is an $n$-element vector representing the desired frequencies of each of the $n$ strategies in a game, and whose elements sum to 1, we compute the Jensen-Shannon (JS) divergence, between $D$ and the distribution resulting from $F_\theta(z)$:

$$JS(D,F_\theta(z)) = \frac{1}{2}KLD(D||M) + \frac{1}{2}KLD(M||F_\theta(z))$$

where $M = \frac{1}{2}(D + F_\theta(z))$ is the average of the two distributions, and KLD is the discrete Kullback–Leibler divergence calculated as:

$$KLD(P,Q) = \sum\limits_{s \in strategies}{P(s)log\frac{P(s)}{Q(s)}}$$

The important property of the JS divergence for our purposes is that it will be minimized when $D = F_\theta(z)$.  Furthermore, $JSD(D,F_\theta(z))$ is a differentiable function parameterized by $\theta$, so we can compute gradients and use SGD to find values of $\theta$ which minimize it.  We construct mini-batches for SGD by samplying from $z$, and apply the training procedure until approximate convergence.  The result is a network which outputs a distribution very close to $D$.

\begin{algorithm}                     
\caption{Initialization Procedure}          
\label{alg1}                           
\begin{algorithmic}
    \REQUIRE $b$, the batch size, $n$, the dimension of $Z$ and $D$, the desired output distribution
    \WHILE{network has not converged}
        \STATE Draw samples $z \in [0,1]^{b \times n}$ from $Z$
        \STATE Compute $F_\theta(z)$
        \STATE Compute $JSD(D,F_\theta(z))$
        \STATE Compute $\nabla_\theta JSD(D,F_\theta(z))$
        \STATE Update $\theta$ by descending $\nabla_\theta$
    \ENDWHILE
\end{algorithmic}
\end{algorithm}

\subsection{Simulating Quasi-Pure Strategies}

To this point, the model as described has been free to map an individual sample $z$ to an arbitrary mixed strategy.  However, it is sometimes desirable to examine models in which an individual must adopt a single pure strategy.  This presents a potential difficulty, as our training and initialization algorithms require that network outputs be differentiable and present meaningful gradients.  Simply clamping the results to 0 or 1 to enforce pure strategies would violate these constraints.  The standard softmax operator is also insufficient, as it does not restrict the outputs from falling anywhere along the [0,1] range. A sigmoid or softmax operation with a high exponent can force the outputs towards 0 or 1, but will present near-zero gradients as they approximate a step function.

\begin{figure}
\centering

\includegraphics[scale=0.75]{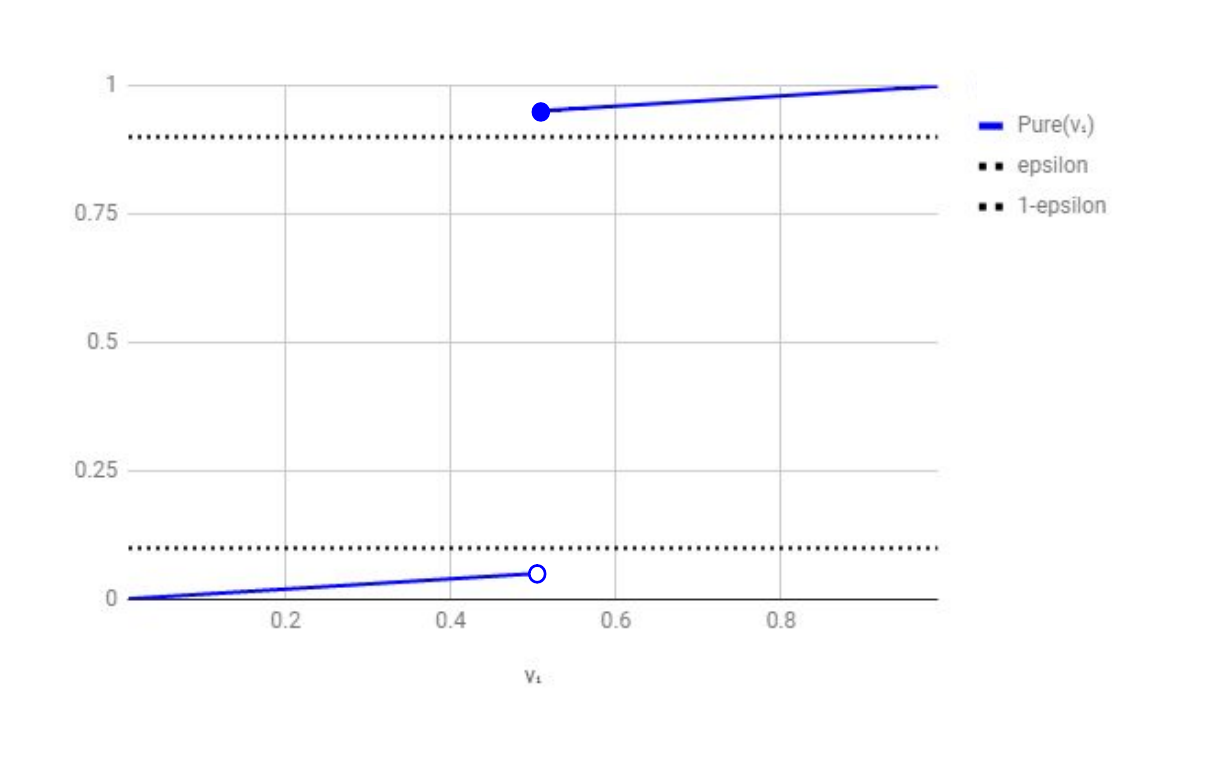}

\caption{$Pure_\epsilon(v_1)$ with $\epsilon = 0.1$ and $max_v = 0.5$.  Note that regardless of the value of $max_v$, $Pure_\epsilon(v_1)$ is within $\epsilon$ of 0 or 1.}
\label{fig:PureGraph}
\end{figure}

To address this, we introduce a soft clamp operator, which we term $Pure_\epsilon(\cdot)$.  This operator takes as input a vector representing a mixed strategy, and maps the largest strategy to a value close to one, and the other strategies to values close to zero.  It does so in a way which preserves the property that the strategy values sum to one, and which presents consistent, meaningful gradients for training.  The operator is parameterized by $\epsilon$, a small value which determines how close to 0 to 1 the outputs must be.  Given a mixed strategy vector $v$, with elements $v_0$ through $v_n$, this operator is defined as:

\[
Pure_\epsilon(v_i) =
\begin{cases}
(1-\epsilon) + \epsilon * v_i & \text{if } \arg\max_v = i\\
\epsilon * v_i & \text{otherwise}
\end{cases}
\]

This operator maps the largest strategy to the range $[1-\epsilon,1]$, and all other strategies to the range $[0,\epsilon]$, creating an output which is very close to a pure strategy.  Further, the derivative of this function is a constant $\epsilon$, allowing the training procedure to direct the network towards improvement.  Figure~\ref{fig:PureGraph} shows the behavior of the $Pure_\epsilon(\cdot)$ operator.

\section{Experiments}

We present two experiments to demonstrate the ability of a neural population model to capture the dynamics predicted by standard methods such as the replicator equation.  First, we will analyze its behavior on the Hawk-Dove game \citep{smith1988evolution}.  This game has been used in the past as a benchmark to analyze the ability of a proposed simulation model to discover and maintain an evolutionary stable state (ESS) \citep{ficici2000effects}.  The Hawk-Dove game has a single ESS, and the dynamics leading towards it are straightforward.  We should expect to see smooth convergence.  The second experiment will focus on the ability of the neural model to capture non-convergent dynamics.  We will use the noisy iterated prisoner's dilemma game, as described in \citep{lindgren1992evolutionary}.  We will restrict ourselves to the dynamics of the four strategies of history length one: All-C, Tit-for-Tat, Anti-Tit-for-Tat, and All-D.  The replicator dynamics of this game result in a continuously changing population distribution, which does not fall into a stable ESS or a simple cycle.

We will compare the behavior of these systems to the behavior shown under the time-discrete replicator equation:
$$x_i(t) = x_i(t-1) + \alpha[f_i(x(t)) - \phi(x(t))]$$
Where $x_i(t)$ is the frequency of strategy $i$ at time $t$, $f_i(\cdot)$ is the fitness of strategy $i$ given a population, and $\phi(\cdot)$ is the average fitness of a population and $\alpha$ is a (small) step size.  We elect to use the time-discrete equation to match the discrete nature of SGD training for the neural model.

\subsection{Model Hyperparameters}

Neural network models are often sensitive to the settings of various hyperparameters.  The details of these parameters and their values are somewhat tangential to the motivation of this work, so we will give only a brief overview of our settings.  The hyperparameters of our model are: the dimension of the latent space, $Z$, the optimizer, batch size, and learning rate used for training, and the value of $\epsilon$ in the $Pure_\epsilon(\cdot)$ layer.  For our experiments, we use the following values.  $Z$ is of dimension 10, networks are trained using the Adam optimizer \citep{kingma2014adam}\footnote{The Adam optimizer is chosen because it is the standard optimizer used in GAN training \citep{chintala2016train}.} with a learning rate of $0.0002$ and a batch size of 2048, and $\epsilon = 0.1$.  In general, these parameters have been chosen through experimentation.

\subsection{Hawks and Doves}

\begin{figure}
\centering

\includegraphics[scale=0.5]{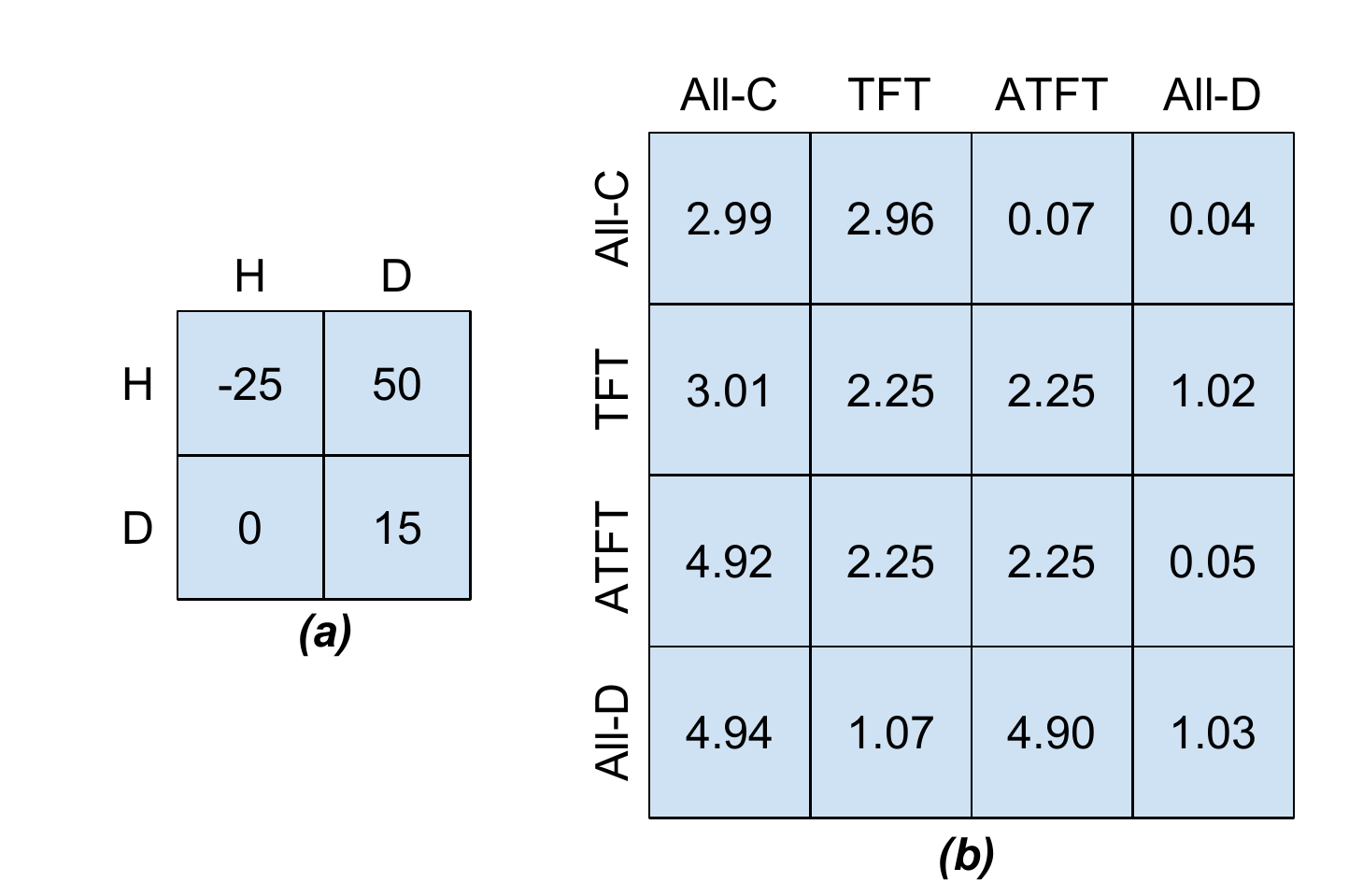}

\caption{Payoff matrices for the hawk-dove game (a) and noisy iterated prisoners' dilemma (b).}
\label{fig:HawkDoveMatrix}
\end{figure}

There are a variety of formulations of the Hawk Dove game in the literature which vary the relative payoffs received by the strategies.  The fundamental dynamic is that when a hawk faces a dove, the hawk receives the highest payoff while the dove receives a small payoff (because the hawk forces the dove away), when two hawks face each other, both receive the lowest payoff (because they fight over the reward, resulting in injury), and when two doves face each other they receive a moderate payoff (because they split the reward without fighting).  We will use the payoffs defined in \citep{ficici2000effects}\footnote{Ficici et al. scale payoffs upward by 26 to make them all positive}, shown in Figure~\ref{fig:HawkDoveMatrix}.  This game as a single ESS in which the population is $\frac{7}{12}$ Hawks.  Any initial mix of Hawks and Doves will converge to this ESS.


We will test the neural model after initialization to a range of starting population ratios.  We will present results for the 10-dimensional $Z$ space, as well as results with a 2-dimensional $Z$ space to allow visualization of how the network maps inputs to strategies.  In order to calculate population frequencies in the neural model, we will draw 10,000 samples from $Z$.






\begin{figure*}
\centering
\includegraphics[scale=0.66]{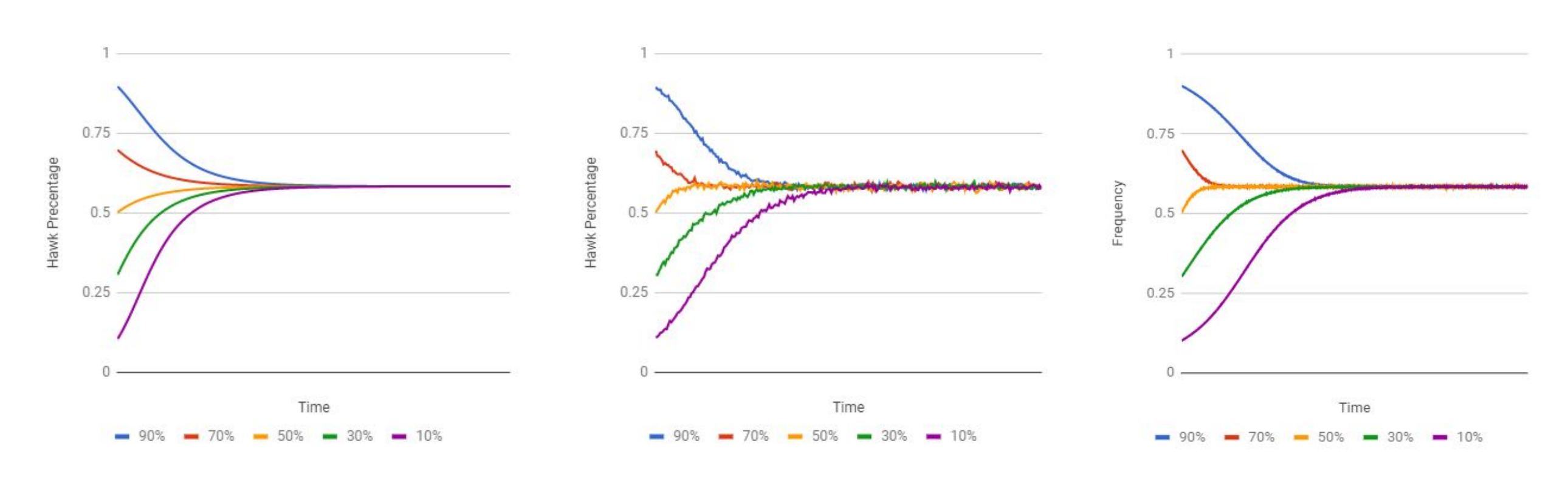}

\caption{Trajectories of the Hawk-Dove game under the replicator equation (left) and neural population model with quasi-pure strategies (center) and mixed strategies (right).  Each graph shows the portion of hawks in the population over time for five different initial population mixes.}
\label{fig:HawkDoveDynamics}
\end{figure*}

\begin{figure}
\centering
\includegraphics[scale=0.7]{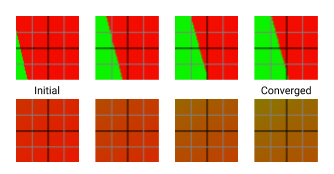}

\caption{Mapping of a 2-dimensional $Z$ to strategies for the Hawk-Dove game.  The top row shows a simulation with quasi-pure strategies, while the bottom row shows a simulation with mixed strategies.  Red represents hawks, and green represents doves.  Intermediate colors represent mixed strategies.}
\label{fig:HawkDovePlotsGrid}
\end{figure}

Figure~\ref{fig:HawkDoveDynamics} shows the trajectory for the game under the replicator equation and the neural population model with both quasi-pure strategies and mixed strategies.  Note that the time axes have been scaled to show similar slopes because the choice of step size is somewhat arbitrary for both models. The dynamics of the neural model are noisier, due to randomness in batch sampling and the indirect nature of the relationship between fitness values and model updates.  Despite this, the model is able to converge to the ESS, and remains near it with only minor fluctuations.  The choice of pure or mixed strategies does not affect the overall dynamics of the model.

Figure~\ref{fig:HawkDovePlotsGrid} presents church-window plots \citep{warde201611} which show the way the network transforms input vectors into strategies.  To enable easy visualization, we restrict the network to a 2-dimensional $Z$ space.  Because the Hawk and Dove game presents simple dynamics, this has little effect on the behavior of the model.  In these plots, the x and y coordinates correspond to the values of a sampled $z$ vector, and the color indicates which strategy the network transforms that vector to, with the red channel representing Hawks, and the green channel representing Doves.  We can see that the mixed strategy network tends to produce a relatively homogeneous population, with only minor spatial variation in strategy (at convergence, the upper left, $z=(0.0,1.0)$, gives about 53\% Hawks, while the lower right, $z=(1.0,0.0)$ gives about 68\% Hawk).  On the other hand, the quasi-pure network creates a strongly differentiated population with a simple boundary.

\subsection{Iterated Prisoner's Dilemma}

We will use the payoffs for the noisy iterated prisoner's dilemma as defined by \citep{lindgren1992evolutionary}, as shown in Figure~\ref{fig:HawkDoveMatrix}(c).  The matrix reflects the average payoffs for a game between four strategies, all-C, which always cooperates, TFT (tit-for-tat), which copies the opponent's last move, ATFT (anti-tit-for-tat), which plays the opposite of the opponent's last move, and all-D, which always defects.  We imagine the players playing an infinitely iterated game using the standard prisoner's dilemma payoffs, as in \citep{axelrod1987evolution}, with the addition of stochastic noise, which will randomly alter a player's move in a small fraction (0.01) of rounds.  The derivation of the payoff values is given in \citep{lindgren1992evolutionary}.

\begin{figure*}[p]
\centering
\includegraphics[scale=0.75]{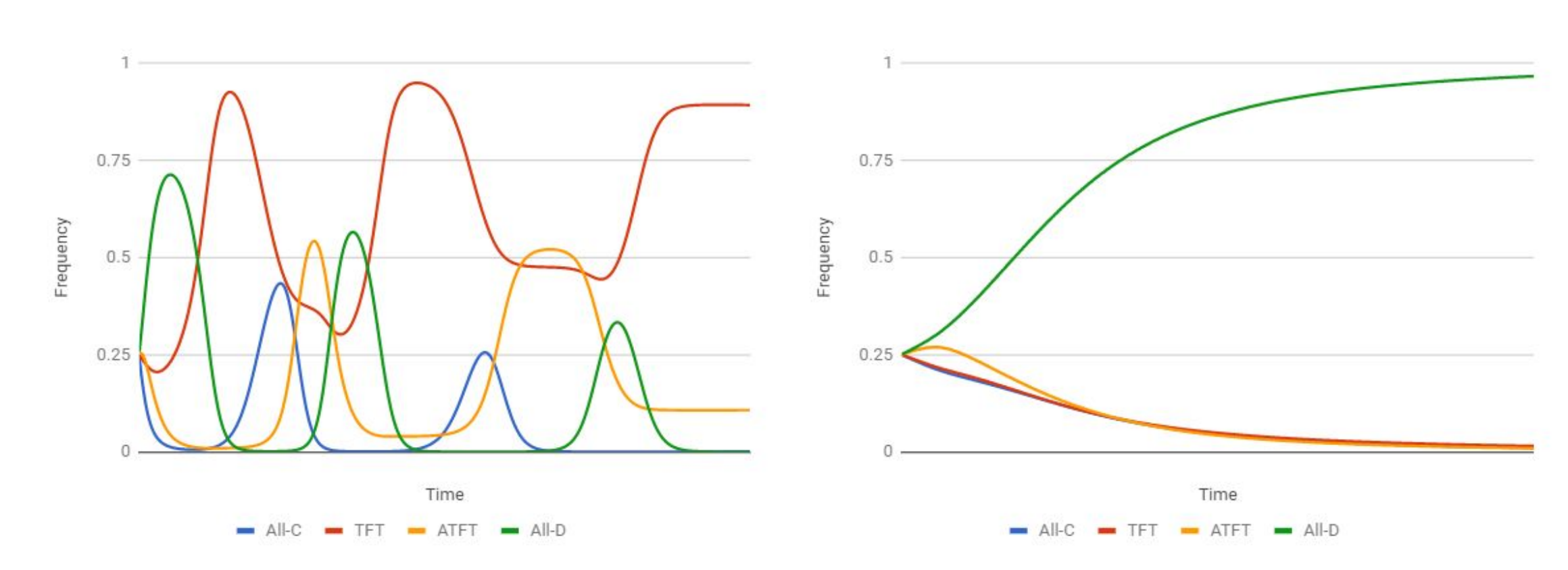}

\caption{Dynamics of the noisy iterated prisoner's dilemma game under the time-discrete replicator equation (left) and the neural model with mixed strategies (right).}
\label{fig:IPD_Replicator}
\end{figure*}

\begin{figure*}[!hp]
\centering
\includegraphics[scale=0.5]{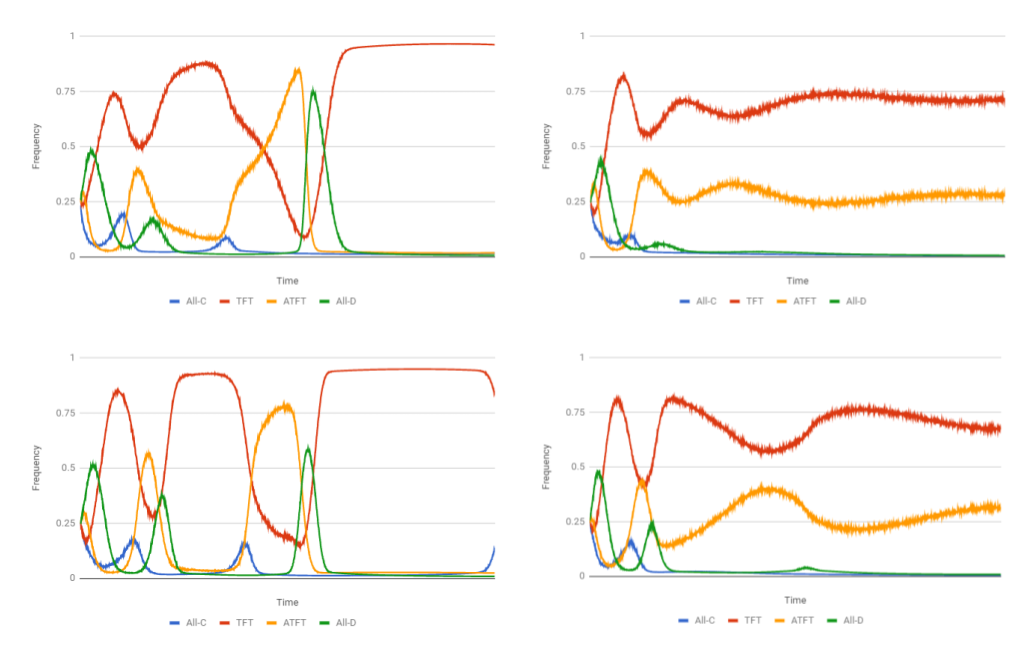}

\caption{Dynamics of the noisy iterated prisoner's dilemma game under the neural population model with quasi-pure strategies.  Four runs are shown, selected to highlight varying degrees of success.}
\label{fig:IPD_Neural_Plots}
\end{figure*}

\begin{figure}
\centering
\includegraphics[scale=0.7]{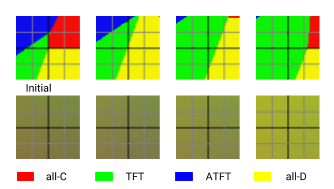}

\caption{Mapping of a 2-dimensional $Z$ to strategies for the noisy iterated prisoner's dilemma .  The top row shows a simulation with quasi-pure strategies, while the bottom row shows a simulation with mixed strategies.}
\label{fig:IPD_Z_Plots}
\end{figure}

Figure~\ref{fig:IPD_Replicator} shows the dynamics of this game as predicted by the time-discrete replicator equation given an initial population which is an equal mix of all strategies.  We see that, unlike the hawk-dove game, the system is non-convergent\footnote{Although the frequencies appear to level off near the end of the simulation, this is actually a transient period of slow change, not a final stable state.}, with strategies rising and falling in frequency in a non-cyclical manner.

As shown in Figure~\ref{fig:IPD_Neural_Plots}, we find that the neural model with quasi-pure strategies is able to approximate these dynamics, in that  the strategies rise and fall in the same order, but the magnitudes and durations of the peaks are often amplified.  As we see in the replicator dynamics, a population consisting almost entirely of TFT and ATFT is semi-stable, and the magnified fluctuations in the neural model sometimes allow it to enter such a state earlier in the simulation, leading to long periods of minimal change.

Unlike in the Hawk-Dove game, we observe a wide variety of outcomes in the noisy iterated prisoner's dilemma, even from identical initial strategy frequencies\footnote{Note that identical initial frequencies does not mean an identical initialization.  A homogeneous population of mixed strategies may have the same total frequencies as a heterogeneous population of pure strategies, and even two homogeneous populations may have different network weights.}.  In some sense, this behavior is expected - the game does not have a simple attractor, and so deviations from the predicted replicator dynamics can be compounded, and steer the system towards a different outcome.  

On the other hand, the mixed strategy model is unable to capture these dynamics, as seen on the right side of Figure~\ref{fig:IPD_Neural_Plots}.  The population converges steadily towards all-D, with only a slight rise in ATFT in the early phases.  

Figure~\ref{fig:IPD_Z_Plots} shows the mapping of a 2-dimensional $Z$ space to strategies for both the quasi-pure and mixed models.  As in the Hawk-Dove game, the mixed model produces a relatively uniform population which smoothly converges towards its final outcome.  By contrast, the quasi-pure model divides the $Z$ space into discrete regions for each strategy, which grow and shrink and as the population distribution changes.  

\section{Discussion}

Our motivation is not to present these neural models as a practical way to investigate games as simple as Hawks and Doves, but rather to demonstrate the applicability of lessons from evolutionary game theory and coevolution to understanding the dynamics of neural network systems in which loss is calculated through interaction. Ultimately, the goal of systems like GANs is to discover a Nash equilibrium using optimization tools designed for finding static minima.  Our neural population models allow us to examine the suitability of these tools to this new application in a setting where we have a firm theoretical understanding of the target dynamics. 

For example, even in the more successful runs, the neural model struggles to maintain the dynamics of the noisy iterated prisoner's dilemma game.  As time progresses, the model swings more and more wildly towards extremes, until it enters long periods of stability in which the activations of the neural network are saturated (which in turn causes vanishingly small gradients, contributing to the stability).  As we see in some of the less successful runs, this dynamic threatens to derail the simulation by prematurely removing diversity from the population (in particular by reducing the population to contain only TFT and ATFT).  These problems seem to mirror problems observed in GAN training, called ``mode collapse'' - the generator often prematurely collapses to produce only a few types of outputs, failing to capture the diversity of the target distribution.  

The very fact that there are successful and unsuccessful runs is itself a cause for concern, the only difference between these runs is the random initialization of the network and the randomly sampled $z$ values used during training.  GAN training is similarly sensitive to initial conditions \citep{lucic2017gans} (as is back-propagation in general \citep{kolen1991back}), but the problem space of image generation is far too high dimensional to allow careful analysis of the effect of parameter initialization.  We propose that the low-dimensionality of neural population models makes them more amenable to detailed understanding of the effects of different initialization procedures and hyperparameter settings.

The failure of the mixed strategy model to capture complex dynamics is somewhat surprising - why is the $Pure_\epsilon(\cdot)$ operator so critical to the system?  Our analysis indicates that this occurs because the network in the mixed strategy model has a tendency to couple the frequencies of different values by using the same internal weights to control multiple outputs.  In particular, in the replicator dynamics and the quasi-pure neural dynamics, we see a strong differentiation between the initial trajectories of All-C and TFT.  All-C drops precipitously, while TFT has only a brief initial decline.  By contrast, in the mixed neural dynamics, we see All-C and TFT decline at the same rate, never diverging from each other by more than a fraction of a percentage point.

The initial separation of All-C and TFT is critical to the long-term dynamics of the system.  TFT is able to outperform All-D only when the All-C strategies which All-D preys upon have been eliminated, but sufficient TFT strategies remain (because TFT performs well against itself compared to the performance of All-D against TFT).  If by the time All-C is nearly eliminated, TFT is also nearly eliminated, the TFT players cannot gain enough fitness to outpace the All-D players, and All-D remains the dominant strategy.

To verify that this is indeed what's happening, we examined the gradients of our loss function with respect to the four strategy outputs.  As expected, we observed that the initial gradients for All-C and TFT are different - they suggest that All-C should fall much faster than TFT.  However, when these gradients are propagated back to the internal weights of the network, the actual impact of the updates is to make both fall at a similar rate.  

We conjecture that maintaining a diverse population, as is forced by the $Pure_\epsilon(\cdot)$ operator, is critical to preventing the degenerate behavior observed in the mixed neural model.  The importance of maintaining population diversity in evolutionary algorithms is well-studied \citep{ursem2002diversity}, and Goodfellow et al. report the occasional occurrence of catastrophic loss of diversity in GAN training.

\section{Conclusion \& Future Work}

We have presented a model which combines elements from the fields of deep learning and artificial life to demonstrate the potential for intellectual cross-pollination between these disciplines.  Our model demonstrates the ability of neural networks to simulate population dynamics, and the applicability of evolutionary game theory results to the behavior of these networks.

Our future work will focus on extending this model with the goal of providing a unifying bridge between the two fields, in the manner of \citep{farmer1990rosetta}.  We will seek to analyze obstacles facing GAN training through the lens of evolutionary game theory, and seek to demonstrate the power of an individual neural network to compactly model an entire population of complex agents, as a step towards open-ended evolution.


\footnotesize
\bibliographystyle{apalike}
\bibliography{references}

\end{document}